\documentclass[letterpaper, journal]{IEEEtran}

\usepackage{mathtools}    
\usepackage{amssymb}
\usepackage{amsfonts}

\usepackage{graphicx}
\usepackage[dvipsnames,table]{xcolor}

\usepackage{cite}
\usepackage{url}
\usepackage{colortbl}
\usepackage{threeparttable}
\usepackage{booktabs}
\usepackage{multirow}
\usepackage{array}
\usepackage{tabularx}
\usepackage{placeins}
\usepackage{makecell}

\makeatletter
\newcommand{\ralmultrow}[2]{\multirow{#1}{*}{\centering\arraybackslash #2}}
\makeatother
\usepackage{siunitx}
\newcommand{\tighttable}{\setlength{\tabcolsep}{3pt}\renewcommand{\arraystretch}{1.1}}
\newcolumntype{Y}{>{\centering\arraybackslash}X}

\usepackage[caption=false,font=footnotesize]{subfig}

\usepackage{algorithm}
\usepackage{algorithmic}

\usepackage{textcomp}
\usepackage{verbatim}
\usepackage[utf8]{inputenc}
\usepackage{balance}
\usepackage{float}
\usepackage{listings}
\usepackage[normalem]{ulem}
\usepackage{pifont}
\usepackage{tcolorbox}
\tcbuselibrary{listings,breakable,skins}
\usepackage{lipsum}
\usepackage{longtable}
\usepackage{seqsplit}
\newif\ifshowcomment
\newif\ifnumberrevision
\newif\ifcolorrevision
\newif\ifstrikeremovel

\showcommenttrue
\numberrevisiontrue
\colorrevisiontrue
\strikeremoveltrue

\newcommand{\lingyao}[1]{\ifshowcomment[{{\textcolor{BrickRed}{Lingyao}}}]\fi}

\newcommand{\lingyaotodo}[1]{\ifshowcomment[{{\textcolor{BrickRed}{Lingyao-TODO}}}]\fi}

\title{Multimodal Physiological Assessment of Contact-Rich Physical Human–Robot Interaction Under Varying Environmental Conditions}

\author{Yanyi Chen\textsuperscript{1}, Xi Wang$^{2}$, Min Deng\textsuperscript{1,3}$^{*}$

\thanks{$^{*}$Corresponding author: Min Deng.}
\thanks{\textsuperscript{1}Yanyi Chen and Min Deng are with the Department of Civil, Environmental, and Construction Engineering, Texas Tech University, Lubbock, TX 79409, USA {(e-mail: yanychen@ttu.edu)}. }

\thanks{\textsuperscript{2}Xi Wang is with the Department of Construction Science, Texas A\&M University, College Station, TX 77843, USA  {(e-mail: xiwang@tamu.edu)}. }

\thanks{\textsuperscript{3}Min Deng is also with the Department of Civil and Environmental Engineering, University of Tennessee, Knoxville, TN 37996, USA {(e-mail: mindeng@utk.edu)}. }
}

\begin{document}
\maketitle

\begin{abstract}
Physical human-robot interaction (pHRI) in real-world settings exposes operators to fluctuating environmental conditions during contact-rich tasks. Traditional task-centric evaluations overlook the physiological burdens imposed by these stressors. Therefore, we conducted a multimodal empirical study involving contact-rich tracing tasks under 18 distinct combinations of temperature, acoustic noise, and illuminance. Synchronously, we recorded electrodermal activity (EDA), surface electromyography (sEMG), eye-tracking data, and subjective environmental comfort ratings. Evaluating these physiological signals alongside execution data revealed hidden physiological costs not captured by objective performance. The results revealed that task performance remained stable across all environmental conditions. Autonomic workload, indexed by tonic skin conductance level (SCL), increased with temperature, while physical and cognitive workload were unaffected. Perceived environmental comfort showed no significant association with tracing error or completion time. These findings reveal a compensatory mechanism where operators maintain consistent performance by increasing their physiological effort to suppress thermal discomfort. Such insight motivates the development of physiology-aware control architectures that leverage real-time physiological metrics to reduce operator workload in unstructured environments.

\end{abstract}
\section{Introduction}
\label{sec:introduction}

\IEEEPARstart {W}{ith} the continued advancement of robotic systems and their transition beyond isolated automation, close collaboration between humans and robots has become increasingly important in domains such as advanced manufacturing \cite{polish2025}, intelligent construction \cite{deng2026integrating, chen2026perception}, and medical surgery\cite{fu2025}. In these settings, robots excel at repetitive, physically demanding, and high-precision tasks, whereas humans remain essential for complex decision-making, adaptation to unexpected situations, and handling uncertainty in dynamic environments. Bridging these complementary strengths, physical human–robot interaction (pHRI) facilitates intuitive physical collaboration by integrating human adaptability and situational awareness with robotic strength, repeatability, and precision \cite{Ogenyi2021,chen2025}.

Among various pHRI applications, contact-rich tasks are particularly challenging. Typical examples include robotic polishing of complex surfaces in advanced manufacturing \cite{polish2025}, cooperative sawing in intelligent construction \cite{Meng2024_Sawing}, and ultrasound scanning on soft tissues in medical surgery \cite{fu2025}. These contact-rich pHRI tasks demand high operational complexity because the human operator must guide the motion along the surface while simultaneously regulating normal contact force. 

To address these challenges, admittance control provides an intuitive interaction framework. It maps measured human-applied forces into robot motion references, enabling compliant physical interaction during human guidance \cite{keeminkAdmittanceControlPhysical2018}. Building on this framework, recent pHRI research has explored adaptive admittance \cite{ferragutiVariableAdmittanceControl2019}, shared control \cite{han2025}, and human state modeling \cite{cre2020} to improve interaction stability, safety, and task performance.
 
However, most existing approaches primarily evaluate interaction quality through external task metrics such as trajectory accuracy, force regulation, or completion time. Such task-centric evaluation neglects the human's internal state, overlooking the operator's underlying physical and cognitive demands, as well as their real-time reactions during interaction. Prior work has shown that multimodal human-state measurements, including electroencephalography (EEG), surface electromyography (sEMG), electrodermal activity (EDA), eye activity, and body-motion cues, can provide complementary information about physical effort, stress, trust, fatigue, attentional demand, engagement, and perceived workload during HRI \cite{Wang2024},\cite{belkaidMutualGazeRobot2021}. Crucially, these studies further suggest that human behavioral and neural responses often modulate to compensate for interaction demands before such effects become fully visible in coarse task-level performance metrics. These findings motivate the need for multimodal approaches that examine not only how accurately a task is completed externally, but also how the interaction is internally experienced and regulated by the human collaborator.

Beyond the interaction design itself, real-world workspaces expose operators to varying environmental conditions, including temperature, illumination, and noise, that can fluctuate over time and differ across sites, influencing well-being, interaction behavior, and collaborative performance \cite{Storm2022}. Prior research indicates that such conditions can lead to visual discomfort, cognitive load, physiological arousal, and fatigue \cite{lan2010effects, DENG2021108098}. In contact-rich pHRI, these effects may further alter how operators regulate interaction forces, allocate attention, and respond to robotic assistance. Despite advances in robot control algorithms for pHRI, how environmental conditions reshape multimodal operator state during contact-rich physical interaction remains insufficiently explored.

In order to overcome these research gaps, we conducted an exploratory multimodal pilot study ($N=12$) where participants performed contact-rich tracing tasks on an unknown corrugated surface. The experiment employed a within-subject design across a factorial combination of thermal, acoustic, and illuminance levels. Across all experimental conditions, the robot operated under an orthogonal shared control method. This method allowed operators to guide the planar motion while the robot autonomously regulated the normal contact force. To capture the operators' multidimensional states, we synchronously recorded three categories of data: objective robotic metrics such as kinematics and interaction wrenches; physiological and behavioral signals including sEMG, EDA, and eye tracking; and subjective environmental comfort ratings. This study addresses three research questions: \textbf{RQ1:} How do thermal, acoustic, and illuminance conditions affect task performance and multimodal operator workload in contact-rich pHRI? \textbf{RQ2:} Which environmental factor exerts the dominant effect on operator physiological responses during contact-rich physical interaction? \textbf{RQ3:} Does perceived environmental comfort correspond to changes in objective task performance?

The contributions of this work include: 1) a multimodal empirical study that decomposes operator workload into physical, cognitive, and autonomic dimensions during contact-rich pHRI under varying environmental conditions; and 2) empirical evidence that task performance remains stable across environmental conditions while autonomic workload increases with temperature, indicating that task performance alone does not fully reflect operator state in real-world workspaces.

\section{Materials and Methods}
\label{sec:methods}

\subsection{Study Design and Protocol}
\label{sec:methods_design}

\begin{figure}[!b]
  \centering
  \includegraphics[width=\columnwidth]{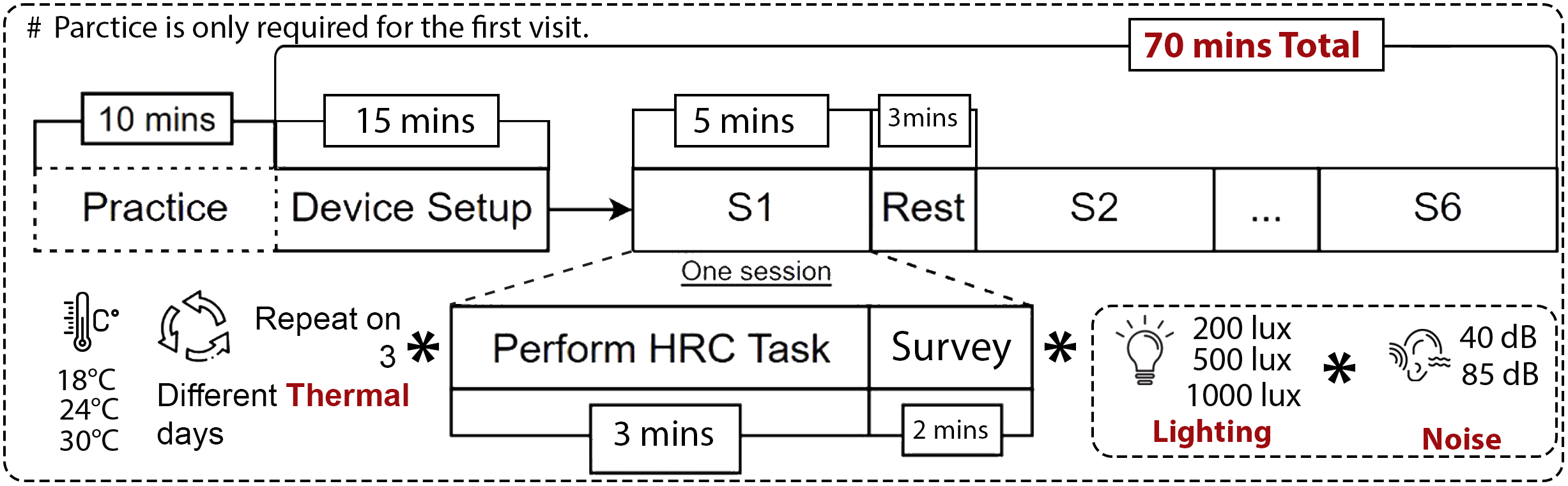}
  \caption{Experimental protocol overview.}
  \label{fig:protocol}
\end{figure}

Figure~\ref{fig:protocol} illustrates the overall experimental protocol, including the timeline and the full factorial design matrix. The study utilized a $3\times2\times3$ within-subject pilot design. To allow participants to adjust to the thermal conditions, the experiment was conducted across three separate laboratory visits on different days, with one specific thermal level (\SIlist{18;24;30}{\celsius}) assigned per visit. These thermal levels were set based on industry standards (i.e., ASHRAE) to represent cool, neutral, and warm environments, respectively, while the relative humidity was maintained at around 45\% \cite{ashrae2017}. 

Within a single visit, the participant completed six consecutive sessions, covering all combinations of the illuminance (\SIlist{200;500;1000}{lux}) lux, with a color temperature of 5000 K, and acoustic noise (\SIlist{40;85}{dB}) conditions. The illuminance levels were determined based on prior human factors research to simulate visual environments suitable for basic physical tasks, standard office work, and precision drawing activities~\cite{DENG2021108098}. Moreover, the selected acoustic noise levels represent a natural quiet baseline and a highly noisy industrial environment threshold, aligning with the occupational noise exposure guidelines recommended by NIOSH~\cite{chan1998occupational}.

The first visit included a \SI{10}{min} practice period to familiarize participants with the system. Each of the three visits began with a \SI{15}{min} device setup and signal-quality check, followed by the six \SI{5}{min} sessions separated by \SI{3}{min} inter-session rests. Each session comprised a \SI{3}{min} contact-rich tracing task and a \SI{2}{min} period for post-trial subjective questionnaires, yielding 18 tasks per participant in total.

\subsection{Participants}
\label{sec:participants}

A total of twelve healthy human subjects ($N=12$; mean age $25.3 \pm 5.3$; 7 female, 5 male) participated in our study. Participants were recruited from the university community and comprised undergraduate, master's, and doctoral students without experience with collaborative robotics. The inclusion criteria required participants to be capable of safely performing light physical activities. Individuals with a history of musculoskeletal, neurological, cardiovascular, or severe respiratory conditions that could compromise force-based interaction, as well as those with skin sensitivities at the planned sEMG and EDA electrode placement sites, were excluded. 

The experimental protocol was approved by the Institutional Review Board (IRB) of Texas Tech University. All participants provided written informed consent prior to data collection and retained the right to withdraw from the study at any time.

\begin{figure*}[!t]
  \centering
  \includegraphics[width=\textwidth]{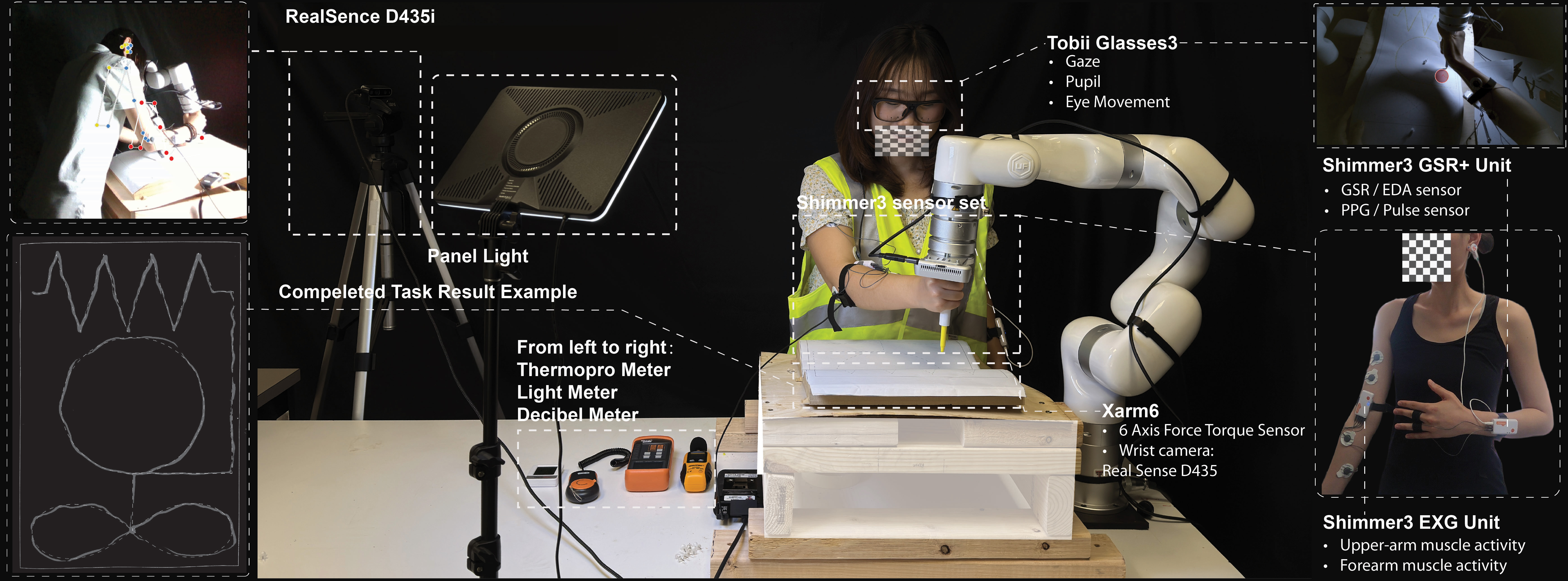}
  \caption{Multimodal contact-rich pHRI apparatus: xArm6 with force-torque sensing, Tobii Glasses~3, Shimmer3 EXG/GSR+, environmental meters, panel lighting, and RealSense D435i.}
  \label{fig:setup}
\end{figure*}

\subsection{Apparatus and Environment control}
\label{sec:platform}

As illustrated in Fig.~\ref{fig:setup}, the collaborative task was performed using a 6-DOF UFactory xArm6 manipulator equipped with a wrist-mounted 6-axis force-torque sensor to capture interaction wrenches. To synchronously capture the operators' psychophysiological and behavioral states, participants wore a Tobii Pro Glasses~3 eye tracker and two wireless biosignal modules: a Shimmer3 EXG unit on the right arm for sEMG, and a Shimmer3 GSR+ unit on the left arm for EDA. Furthermore, a tripod-mounted Intel RealSense D435i RGB-D camera was deployed to record upper-body kinematics. To ensure operator safety under the shared control architecture, the robot was operated at safe kinematic scale, and an emergency stop switch was positioned within immediate reach of both the participant and the experimenter.

To ensure strict environmental regulation, all tasks were conducted in a sealed laboratory room with the central HVAC system disabled. Thermal conditions were actively regulated using portable heating and cooling devices. To standardize intrinsic thermal insulation across all participants in accordance with ASHRAE thermal comfort guidelines, all subjects were required to wear long trousers and short-sleeved shirts (yielding an approximate clothing insulation of \SI{0.5}{clo}). Illuminance levels were configured using an adjustable LED panel, while blackout curtains were tightly drawn to prevent daylight intrusion. For the acoustic environment, the natural baseline room noise was approximately \SI{40}{dB}, whereas the high-noise condition was simulated by broadcasting industrial white noise at \SI{85}{dB} via acoustic playback equipment.

To guarantee consistency with the intended factorial design, environmental parameters were continuously monitored using industrial-grade meters placed relative to the operator. Temperature was measured at the abdomen level ($\pm\SI{0.5}{\celsius}$), acoustic noise was verified near ear height ($\pm\SI{5}{dB}$), and illuminance was tracked at the center of the workspace ($\pm\SI{30}{lux}$).

\subsection{Contact-Rich pHRI Tracing Task}
\label{sec:task}

\begin{figure}[!b]
  \centering
  \includegraphics[width=\linewidth]{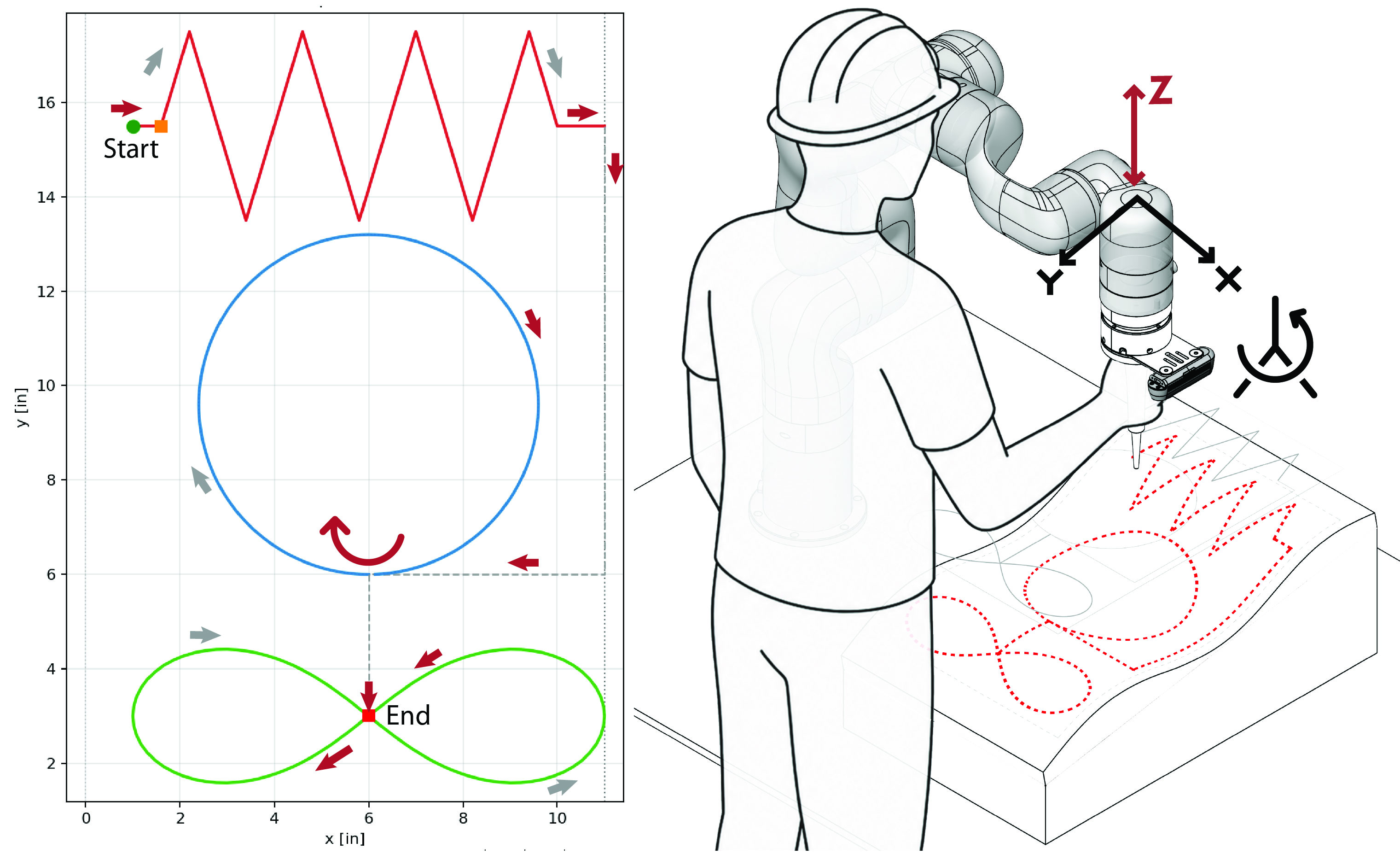}
  \caption{Contact-rich tracing task: nominal four-segment path, motion direction, and an example scene on the inclined corrugated surface.}
  \label{fig:task_traj}
\end{figure}

Participants grasped a custom end-effector to physically guide a marker along a predefined reference path on an inclined, corrugated surface as illustrated in Fig.~\ref{fig:task_traj}. They were instructed to trace the full path accurately and smoothly while maintaining continuous marker-surface contact.

As Shown in Fig.~\ref{fig:task_traj}, the task path concatenated four representative geometric segments in sequence: a zigzag, a long straight line, a circle, and a Bernoulli lemniscate. These segments were specifically designed to elicit diverse human interaction demands, encompassing sharp turns, steady linear motion, continuous curvature, and self-crossing geometry. For the contact board, it is featured a mild sinusoidal corrugation along the primary move direction and was mounted at an approximately \SI{10}{\degree} inclination relative to the horizontal plane.

\subsection{Orthogonal Shared Control}
\label{sec:control}
Inspired by hybrid position/force control, the task space is orthogonally decoupled into a human-guided subspace (planar translation $X$--$Y$ and rotation) and a robot-regulated normal subspace ($Z$)~\cite{ferragutiVariableAdmittanceControl2019}. The control parameters introduced in this section, including the planar deadband bounds and the velocity limits, were empirically selected and tuned through extensive pilot testing based on system performance and recent studies during interaction tasks.

The robot operates in Cartesian velocity mode under a zero-stiffness ($k_v = 0$) admittance controller at \SI{500}{Hz}:
\begin{equation}
    \mathbf{M}_v \dot{\mathbf{v}}_d + \mathbf{B}_v \mathbf{v}_d = \mathbf{u},
    \label{eq:virtual_dynamics}
\end{equation}
where $\mathbf{v}_d\in\mathbb{R}^6$ is the velocity command, and
$\mathbf{M}_v$, $\mathbf{B}_v$ are diagonal virtual inertia and damping matrices
configured as $\mathbf{M}_v=\mathrm{diag}(1.0,1.0,1.0,0.15,0.15,0.15)$ and $\mathbf{B}_v=\mathrm{diag}(120,120,450,4.5,4.5,4.5)$.  Furthermore, to suppress oscillations during direction changes, the effective damping $B_{v,i}$ was scaled by a reversal damping factor $\beta=\num{1.8}$ whenever the commanded input opposed the current velocity along axis $i$ (i.e., $u_i v_{d,i} < 0$). The admittance input $\mathbf{u}=[\mathbf{u}_{xy}^{\top}, u_z, \mathbf{u}_{\tau}^{\top}]^{\top}\in\mathbb{R}^6$ is synthesized channel-wise from the measured external wrench $\mathbf{W}_{\mathrm{ext}}=[\mathbf{F}_{\mathrm{ext}}^{\top}, \boldsymbol{\tau}_{\mathrm{ext}}^{\top}]^{\top}$.

In the planar motion subspace, a deadband with linear transition shaping maps
the applied planar force $f_{xy}=\|\mathbf{F}_{\mathrm{ext},xy}\|$ to the
velocity command $\mathbf{u}_{xy}$:
\begin{equation}
    \mathbf{u}_{xy} =
    \begin{cases}
        \mathbf{0}, & f_{xy} \le F_{\mathrm{dz}}, \\[4pt]
        G_{xy}\,\mathbf{F}_{\mathrm{ext},xy}\,\dfrac{f_{xy}-F_{\mathrm{dz}}}{F_{\mathrm{dz}}^{U}-F_{\mathrm{dz}}},
        & F_{\mathrm{dz}} < f_{xy} < F_{\mathrm{dz}}^{U}, \\[8pt]
        G_{xy}\,\mathbf{F}_{\mathrm{ext},xy}, & f_{xy} \ge F_{\mathrm{dz}}^{U},
    \end{cases}
    \label{eq:planar_input}
\end{equation}
where $G_{xy}=\num{0.55}$ is the admittance gain, $F_{\mathrm{dz}}=\SI{2.8}{N}$
and $F_{\mathrm{dz}}^{U}=\SI{3.6}{N}$ are the lower and upper deadband
thresholds for suppressing sensor noise and unintended tremor, and the linear
transition between them ensures smooth velocity scaling at motion onset.
The resulting $X$ and $Y$ velocity  were limited to $\pm v_{\max}$ ($\SI{38}{mm/s}$).

In the normal force subspace, the robot executes an autonomous contact-force
tracking task against the unknown corrugated surface, with
tracking error $e_f = F_{\mathrm{ext},z} - F_d$ defined relative to the desired
contact force $F_d = \SI{5}{N}$. The normal command $u_z$ is given by:
\begin{equation}
    u_z =
    \begin{cases}
        0, & |F_{\mathrm{ext},z}| < F_{\mathrm{c}} \ \lor \ |e_f| \le \epsilon_f, \\[4pt]
        \alpha_z\,G_z\left(e_f - \epsilon_f\right), & e_f > \epsilon_f, \\[4pt]
        G_z\left(e_f + \epsilon_f\right), & e_f < -\epsilon_f,
    \end{cases}
    \label{eq:normal_input}
\end{equation}
where $G_z=\num{0.38}$ is the baseline error gain, $F_{\mathrm{c}}=\SI{0.35}{N}$
and $\epsilon_f=\SI{0.5}{N}$ are the contact activation threshold and force error
deadband, respectively, and $\alpha_z=\num{0.40}$ is an attenuation factor applied exclusively during upward retreat ($e_f > \epsilon_f$) to prevent abrupt lift-off.
The $Z$ velocity was limited at $v_{\max}$ for downward moving and at the lower $v_{\mathrm{up}}=\SI{15}{mm/s}$ for upward retreat.

In the rotational subspace, human-applied torque $\tau=\|\boldsymbol{\tau}_{\mathrm{ext}}\|$ drives end-effector rotation above a deadband threshold:
\begin{equation}
    \mathbf{u}_{\tau} =
    \begin{cases}
        \mathbf{0}, & \tau < \tau_{\mathrm{th}}, \\[4pt]
        G_{\tau}\,\boldsymbol{\tau}_{\mathrm{ext}}, & \tau \ge \tau_{\mathrm{th}},
    \end{cases}
    \label{eq:rot_input}
\end{equation}
with gain $G_{\tau}=\num{0.28}$ and deadband threshold $\tau_{\mathrm{th}}=\SI{0.1}{N\,m}$.
Rotational velocity were limited to $\pm\omega_{\max}=\SI{22}{deg/s}$.

\subsection{Outcome Measures}
\label{sec:metric}

During the experiments, robot kinematics, interaction wrenches, and admittance states were recorded at the \SI{500}{Hz} control rate.
Human physiological and behavioral signals were acquired in parallel at their sampling rates: forearm and upper-arm sEMG at \SI{512}{Hz} with the Shimmer3 EXG unit, GSR conductance at \SI{256}{Hz} with the Shimmer3 GSR+, binocular pupil diameter at \SI{100}{Hz}, and RGB video at \SI{30}{fps} for image capture. After each trial, participants rated environmental comfort on a 7-point Likert scale.

Recent pHRI assessments frequently characterize operator workload along a 
single dimension or focus primarily on objective task performance indicators such as execution time and accuracy. However, established instruments such as the NASA Task Load Index explicitly distinguish physical from mental demand, reflecting the multidimensional nature of workload~\cite{HART1988139}. Meanwhile, these dimensions are not always correlated: prior work has shown that reductions in physical demand do not necessarily reduce cognitive or psychological burden~\cite{la2023,mol2026}.

Accordingly, we evaluate objective \textit{Task Performance} using the commonly adopted metrics of tracing error and completion time. Beyond task execution, we explicitly decompose operator workload into three measurable physiological dimensions: \textit{Physical Workload}, \textit{Cognitive Workload}, and \textit{Autonomic Workload}. Table~\ref{tab:metrics_mapping} provides the full mapping of outcome metrics, summary statistics, and raw data sources.

\subsubsection{\textbf{Task Performance}}
\label{sec:task_performance_metrics}

\emph{Tracing error} was computed as the planar root-mean-square error (RMS) between the executed end-effector Tool Center Point (TCP) trajectory and the pre-defined reference path.
\emph{Completion time} was the duration of the retained active motion window for each trial.

\subsubsection{\textbf{Physical Workload}}

Physical workload was estimated from right-arm sEMG using the energy cost processing method of Meng et al.~\cite{Meng2024_Sawing}. Two bipolar sEMG channels were recorded from the forearm and upper arm, respectively. Each signal was full-wave rectified, smoothed using a 200\,ms RMS window, and normalized to yield the muscle activation level $\rho_j(t)\in[0,1]$ ($j=\mathrm{forearm,\ upper\ arm}$). Cumulative muscular effort was then computed by integrating the sum of the squared activations of both muscles over the trial duration:
\begin{equation}
    E^h=\int_{t_0}^{t_e} \left[ \rho_{\mathrm{forearm}}^2(t) + \rho_{\mathrm{upperarm}}^2(t) \right] dt.
\end{equation}
The trial-averaged $E^h$ served as the physical workload metric, where higher values indicate greater muscular exertion.

\begin{table}[!t]
\centering
\caption{Outcome metrics and data sources.}
\label{tab:metrics_mapping}
\begingroup
\setlength{\tabcolsep}{18pt}
\renewcommand{\arraystretch}{1.15}
\tighttable
\begin{tabular}{@{}ll|c|c@{}}
\toprule
\multicolumn{1}{l}{\makecell[l]{\textbf{Category}}} &
\multicolumn{1}{l}{\makecell[l]{\textbf{Metric}}} &
\multicolumn{1}{c}{\makecell[c]{\textbf{Outcome} \\ \textbf{measure}}} &
\multicolumn{1}{c}{\makecell[c]{\textbf{Raw} \textbf{data}}} \\
\midrule
\ralmultrow{2}{\makecell[l]{Task \\ performance}}
& \makecell[l]{Tracing error} & Planar RMSE & TCP trajectory \\
\cmidrule(lr){2-4}
& \makecell[l]{Completion time} & Task duration & Robot timestamps \\
\midrule
\ralmultrow{3}{Workload}
& \makecell[l]{Physical  workload} & $E^h$ & sEMG \\
\cmidrule(lr){2-4}
& \makecell[l]{Cognitive workload} & $-\mathrm{LHIPA}_z$ & Pupil diameter \\
\cmidrule(lr){2-4}
& \makecell[l]{Autonomic workload} & $\mathrm{SCL}_z$ & GSR conductance \\
\midrule
\ralmultrow{3}{\makecell[l]{Subjective \\ environmental \\ comfort ratings}}
& \makecell[l]{Thermal comfort} & Comfort rating & --- \\
\cmidrule(lr){2-4}
& \makecell[l]{Visual comfort} & Comfort rating & --- \\
\cmidrule(lr){2-4}
& \makecell[l]{Acoustic comfort} & Comfort rating & --- \\
\bottomrule
\end{tabular}
\endgroup
\end{table}
\subsubsection{\textbf{Cognitive Workload}}
The LHIPA was selected to quantify cognitive workload. LHIPA applies wavelet decomposition to separate low-frequency and high-frequency pupillary oscillations, where low-frequency components reflect parasympathetic activity and slow luminance adaptation, and high-frequency components track sympathetic arousal and cognitive effort~\cite{duch2020}. This separation makes LHIPA less sensitive to baseline drift and luminance variation, which is particularly relevant for continuous tracking tasks. As LHIPA is negatively correlated to cognitive load, the raw values were inverted and normalized using within-subject $z$-score standardization. The trial-averaged metric, denoted $-\mathrm{LHIPA}_z$, served as the cognitive workload.

\begin{table*}[!t]
  \centering
  \caption{REPORTED DESCRIPTIVE AND INFERENTIAL STATISTICS OF THE OBJECTIVE MEASURES}
  \label{tab:all_statistics_summary}
  \setlength{\tabcolsep}{4.5pt}
  \renewcommand{\arraystretch}{0.9}
  
  \newcommand{\best}[1]{\textcolor{green!60!black}{\textbf{#1}}} 
  \newcommand{\worst}[1]{\textcolor{red}{\textbf{#1}}}   

  \begin{tabular}{@{}ll | cc | ccc @{}}
    \toprule
    & & \multicolumn{2}{c|}{\textbf{Task Performance}} & \multicolumn{3}{c}{\textbf{Workload}} \\
    \cmidrule(lr){3-4} \cmidrule(l){5-7}
    \textbf{Factor} & \textbf{Level}
    & \makecell{\textbf{Tracing error} \\ ($M \pm SD$)}
    & \makecell{\textbf{Completion time} \\ ($Mdn$, IQR)}
    & \makecell{\textbf{Physical workload} \\ ($Mdn$, IQR)}
    & \makecell{\textbf{Cognitive workload} \\ ($M \pm SD$)}
    & \makecell{\textbf{Autonomic workload} \\ ($M \pm SD$)} \\
    \midrule
    
    \multirow{6}{*}{\textbf{Temperature}} 
    & 18\,$^\circ$C & \best{8.4 $\pm$ 0.9} & 227 (48) & \best{27.8 (14.5)} & \best{-0.30 $\pm$ 0.50} & \best{-0.77 $\pm$ 0.30} \\
    & 24\,$^\circ$C & 8.6 $\pm$ 1.2 & \best{222 (96)} & 36.2 (22.7) & -0.01 $\pm$ 0.58 & \worst{0.51 $\pm$ 0.64} \\
    & 30\,$^\circ$C & \worst{8.7 $\pm$ 1.3} & \worst{238 (62)} & \worst{37.8 (17.0)} & \worst{0.19 $\pm$ 0.45} & 0.45 $\pm$ 0.47 \\
    \cmidrule{2-7}
    & \textit{Main effect} & \makecell[c]{$F(2)=0.14$ \\ $p=0.868$} & \makecell[c]{$\chi^2(2)=4.67$ \\ $p=0.097^\dagger$} & \makecell[c]{$\chi^2(2)=2.36$ \\ $p=0.307$} & \makecell[c]{$F(2)=2.30$ \\ $p=0.120$} & \makecell[c]{\textbf{F(2)=25.96} \\ \textbf{p<0.001$^{**}$}} \\
    & \textit{Effect size} & $\eta^2=0.01$ & $W=0.19$ & $W=0.11$ & $\eta^2=0.15$ & \textbf{$\eta^2=0.61$} \\
    & \textit{Piece-wise} & --- & --- & --- & --- & \makecell[c]{18--24\,$^\circ$C \& 18--30\,$^\circ$C: \textbf{<0.001$^{**}$} \\ 24--30\,$^\circ$C: 0.861} \\
    \midrule

    \multirow{4}{*}{\textbf{Noise}} 
    & 40\,dB & \worst{8.6 $\pm$ 1.0} & \worst{241 (60)} & \best{31.0 (20.1)} & \worst{-0.00 $\pm$ 0.46} & \worst{0.12 $\pm$ 0.26} \\
    & 85\,dB & \best{8.5 $\pm$ 0.9} & \best{240 (71)} & \worst{34.7 (20.4)} & \best{-0.06 $\pm$ 0.26} & \best{-0.01 $\pm$ 0.19} \\
    \cmidrule{2-7}
    & \textit{Main effect} & \makecell[c]{$t=0.80$ \\ $p=0.443$} & \makecell[c]{$t=1.27$ \\ $p=0.232$} & \makecell[c]{$t=-1.30$ \\ $p=0.221$} & \makecell[c]{$t=0.36$ \\ $p=0.725$} & \makecell[c]{$t=0.99$ \\ $p=0.345$} \\
    & \textit{Effect size} & $d=0.23$ & $d=0.37$ & $d=-0.37$ & $d=0.11$ & $d=0.28$ \\
    \midrule

    \multirow{5}{*}{\textbf{Illuminance}} 
    & 200\,lx & 8.5 $\pm$ 1.1 & 231 (60) & \worst{34.2 (21.9)} & \best{-0.07 $\pm$ 0.52} & \best{-0.08 $\pm$ 0.40} \\
    & 500\,lx & \best{8.5 $\pm$ 0.9} & \worst{254 (79)} & 31.8 (11.4) & \worst{0.07 $\pm$ 0.47} & 0.10 $\pm$ 0.17 \\
    & 1000\,lx & \worst{8.6 $\pm$ 0.9} & \best{230 (59)} & \best{26.7 (23.3)} & 0.01 $\pm$ 0.33 & \worst{0.17 $\pm$ 0.30} \\
    \cmidrule{2-7}
    & \textit{Main effect} & \makecell[c]{$F(2)=0.01$ \\ $p=0.990$} & \makecell[c]{$F(2)=0.14$ \\ $p=0.868$} & \makecell[c]{$F(2)=0.12$ \\ $p=0.885$} & \makecell[c]{$F(2)=0.27$ \\ $p=0.769$} & \makecell[c]{$F(2)=2.04$ \\ $p=0.146$} \\
    & \textit{Effect size} & $\eta^2=0.00$ & $\eta^2=0.01$ & $\eta^2=0.01$ & $\eta^2=0.02$ & $\eta^2=0.11$ \\
    \bottomrule
  \end{tabular}
  
  \par\vspace{0.8em}
  \begin{minipage}{\textwidth}
    \footnotesize
    \textit{Note:} $N=12$. \best{Green bold} = best, \worst{Red bold} = worst within each factor and measure.
    Normality of subject-level marginal scores was assessed with Shapiro--Wilk ($\alpha=0.01$); Tracing error, Cognitive workload ($-\mathrm{LHIPA}_z$), Autonomic workload ($\mathrm{SCL}_z$) are presented as $M \pm SD$; Completion time, Physical workload ($E^h$) are presented as $Mdn$ (IQR).
    Effect sizes: Cohen's $d$ (paired $t$), Kendall's $W$ (Friedman), $\eta^2$ (one-way ANOVA).
    Bold values denote statistical significance. $^\dagger\,p<0.10$; $^{*}\,p<0.05$; $^{**}\,p<0.01$ (main effects and Holm-adjusted piece-wise contrasts).
  \end{minipage}
\end{table*}

\subsubsection{\textbf{Autonomic Workload}}
Raw GSR signals were preprocessed following a standard pipeline~\cite{DENG2021108098}: apparent outliers were removed, and a Butterworth low-pass filter (2\,Hz) was applied to attenuate movement artifacts and high-frequency noise. Continuous Decomposition Analysis (CDA) was then applied to separate the signal into a phasic component (SCR) and a tonic component (SCL). As SCL reflects background sympathetic arousal associated with stress, the trial-mean SCL was extracted as the summary statistic. To account for individual differences in baseline conductance, values were normalized using within-subject $z$-score standardization. The trial-averaged $\mathrm{SCL}_z$ served as the autonomic workload metric, where higher values indicate greater autonomic arousal.

\subsection{Statistical Analysis}
\label{sec:analysis}

To isolate the main effect of each environmental factor, marginal means were computed by averaging across the remaining factors for each participant. Normality of these marginal scores was assessed with the Shapiro-Wilk test ($\alpha=0.01$). For normally distributed outcomes, a one-way ANOVA was applied to three-level factors (temperature and illuminance) and a paired $t$-test to the two-level factor (noise); otherwise, Friedman tests and Wilcoxon signed-rank tests were used accordingly. Post-hoc pairwise comparisons were conducted for significant omnibus effects on three-level factors using paired $t$-tests or Wilcoxon signed-rank tests with Holm-Bonferroni correction. Effect sizes were reported as Cohen's $d$ for $t$-tests, Kendall's $W$ for Friedman tests, and $\eta^2$ for ANOVAs. Exploratory Spearman rank correlations were computed at the trial level across participants between comfort ratings and objective performance metrics.

\section{Quantitative Outcomes}
\label{sec:results}

\begin{figure*}[!t]
  \centering
  \includegraphics[width=\textwidth]{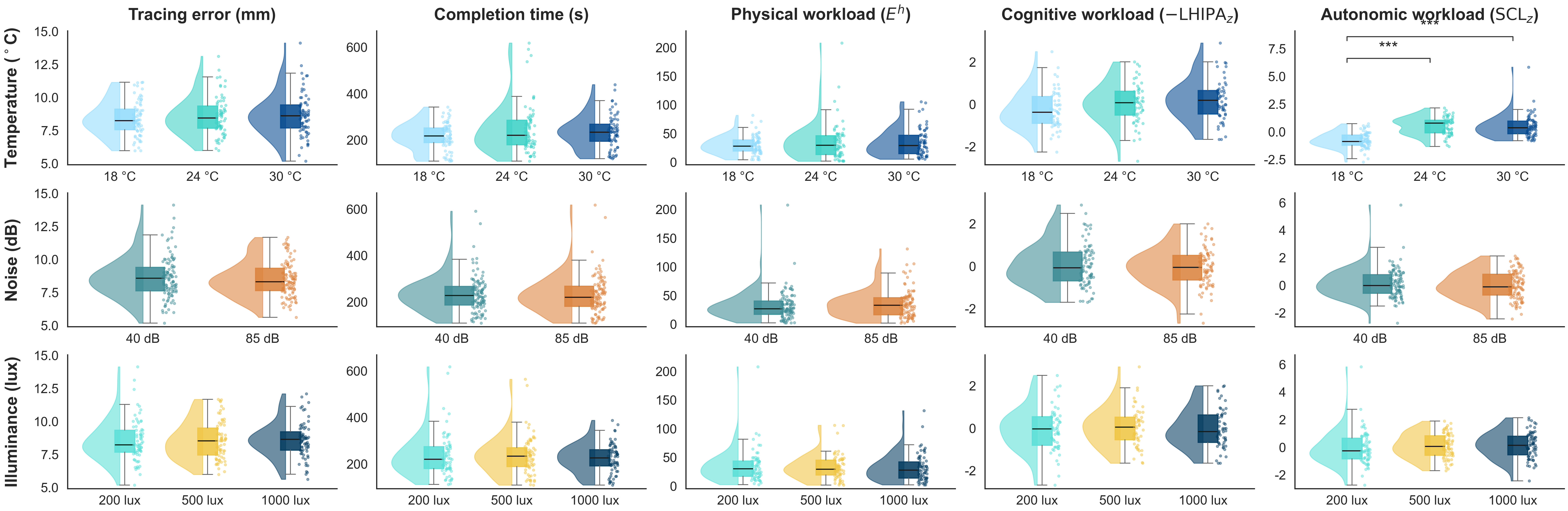}
  \caption{Task performance and workload outcomes by environment factor ($N{=}12$). Statistically significant Holm-adjusted pairwise differences are annotated; $^{**}$: $p<0.01$.}
  \label{fig:env_outcomes_raincloud}
\end{figure*}

\begin{figure}[!b]
  \centering
  \includegraphics[width=\linewidth]{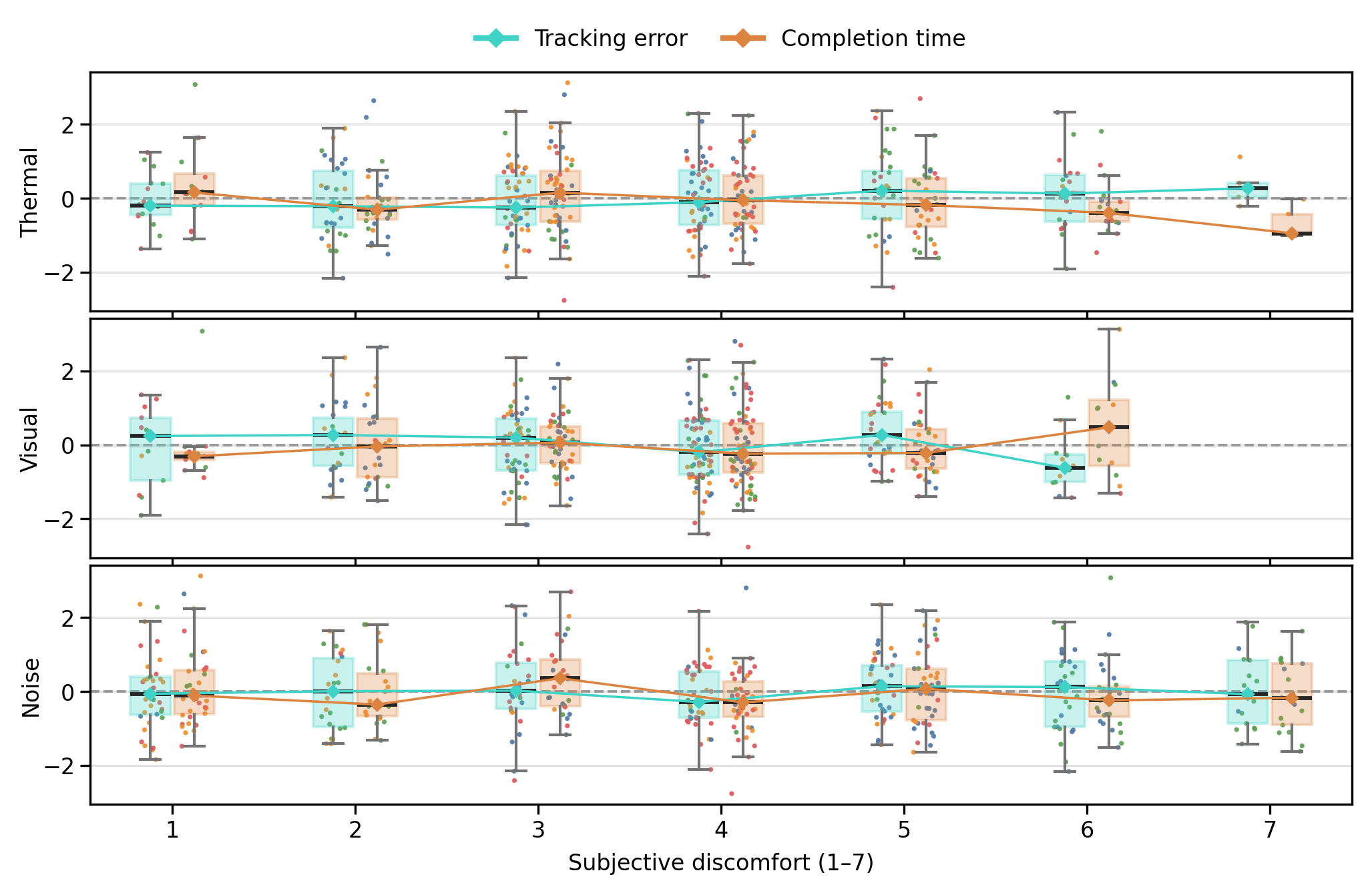}
  \caption{Within-subject $z$-scored tracing error and completion time versus environmental comfort ratings ($N{=}12$).}
  \label{fig:comfort_performance}
\end{figure}

The experimental results are presented below, highlighting the effects of temperature, acoustic noise, and illuminance on task performance, multidimensional workload, and environmental comfort ratings during contact-rich tracing ($N{=}12$). Findings are structured by measure type (Table~\ref{tab:metrics_mapping}), referencing Fig.~\ref{fig:env_outcomes_raincloud}, Fig.~\ref{fig:comfort_performance}, and the inferential evaluation in Table~\ref{tab:all_statistics_summary}.

\textbf{Task Performance} was evaluated using two objective metrics: tracing error and completion time. The columns~1-2 of Fig.~\ref {fig:env_outcomes_raincloud} show the distribution of scores for each environmental factor, with inferential statistics summarized in Table~\ref{tab:all_statistics_summary}.

No significant main effects were found for tracing error across temperature ($F(2)=0.14$, $p=0.868$), noise ($t=0.80$, $p=0.443$), or illuminance ($F(2)=0.01$, $p=0.990$). Cohort means ranged from \SI{8.4}{mm} to \SI{8.7}{mm} across thermal levels. Completion time showed no significant main effects of noise ($t=0.99$, $p=0.345$) or illuminance ($F(2)=0.14$, $p=0.868$). A marginal temperature effect was observed ($\chi^2(2)=4.67$, $p=0.097$; Kendall $W=0.19$), with session medians between \SI{222}{s} and \SI{238}{s} across \SIlist{18;24;30}{\celsius}, but no Holm-adjusted pairwise contrast reached significance.

\textbf{Operator Workload} was evaluated along physical, cognitive, and autonomic dimensions. Columns~3-5 of Fig.~\ref{fig:env_outcomes_raincloud} and Table~\ref{tab:all_statistics_summary} report the distributions and inferential statistics for each workload dimension.

No significant main effects were found for physical or cognitive workload across any environmental factor. Autonomic workload differed significantly across thermal levels ($F(2)=25.96$, $p<0.001$, $\eta^2=0.61$), with mean scores lowest at \SI{18}{\celsius} ($-0.77 \pm 0.30$) and higher at \SI{24}{\celsius} ($0.51 \pm 0.64$) and \SI{30}{\celsius} ($0.45 \pm 0.47$). Holm-adjusted pairwise comparisons indicated lower autonomic workload at \SI{18}{\celsius} than at \SI{24}{\celsius} and \SI{30}{\celsius} (both $p<0.001$); the \SI{24}{\celsius} versus \SI{30}{\celsius} contrast was not significant ($p=0.861$). Noise and illuminance showed no significant main effects on autonomic workload ($t=0.99$, $p=0.345$; $F(2)=2.04$, $p=0.146$).

\textbf{Environmental Comfort Ratings}
were collected on a 7-point Likert scale after each trial along thermal, visual, and acoustic dimensions. Fig.~\ref{fig:comfort_performance} relates these ratings to within-subject $z$-scored tracing error and completion time; comfort ratings were not formally tested against environmental factors.

Tracing error and completion time remained near the within-participant baseline across all comfort-rating levels within each dimension. Trial-level Spearman correlations between comfort ratings and both performance measures were weak ($|\rho|<0.13$; all $p>0.05$).

\section{Interpretation and implication}
\label{sec:discussion}

Based on the inferential statistics of robot kinematic metrics and physiological metrics, this study evaluated the effects of varying environmental conditions on task performance and multimodal operator workload during contact-rich pHRI. Across 18 combinations of temperature, acoustic noise, and illuminance, task performance remained stable while autonomic workload increased selectively with thermal level. This finding suggests that operators maintained execution quality under adverse ambient conditions while sustaining elevated autonomic arousal, which increases mental workload.

\textbf{Task performance remained stable across environmental conditions.} Tracing error showed no significant main effects across temperature, noise, or illuminance, with cohort means ranging narrowly from \SI{8.4}{mm} to \SI{8.7}{mm}. Completion time was unaffected by noise and illuminance. A marginal temperature effect on completion time was observed but did not survive post-hoc correction. These results suggest that the tested environmental conditions did not degrade kinematic execution within this task. However, stable task performance alone does not imply that operators were unaffected by the environment.

\textbf{Temperature elevated autonomic workload without affecting task execution.} Autonomic workload was the only measure to show a significant main effect. Mean SCL$_z$ was lowest at \SI{18}{\celsius} and higher at both \SI{24}{\celsius} and \SI{30}{\celsius}, with both pairwise contrasts reaching significance after Holm adjustment. The \SI{24}{\celsius} versus \SI{30}{\celsius} contrast was not significant, suggesting that autonomic arousal may saturate above a moderate thermal threshold rather than scaling linearly with temperature. Physical and cognitive workload showed no significant effects across any environmental factor. These results indicate that thermal load was reflected in autonomic responses before any degradation in tracing execution.

\textbf{Perceived comfort decoupled from kinematic performance.} Trial-level Spearman correlations between comfort ratings across all environmental dimensions and both task performance metrics were weak. Operators who reported lower environmental comfort did not show correspondingly higher tracing error or longer completion times. Combined with the autonomic workload findings, this decoupling indicates that subjective comfort ratings, physiological measures, and objective performance metrics each capture distinct aspects of operator state. Objective task performance analysis alone is insufficient to characterize operator state in contact-rich pHRI under varying environmental conditions in a real workspace. Therefore, multimodal physiological assessment is necessary to reveal the full human responses for a human-centered pHRI method to reduce hidden operator workload.

\textbf{Limitations.} Several limitations warrant consideration. First, the participant pool was recruited from a university setting and lacked prior pHRI experience; thus, generalizing these findings to skilled industrial operators should be done with caution. Second, this study utilized typical, controlled settings for temperature, lighting, and noise to establish baseline physiological responses. In practice, however, real-world environmental conditions are inherently broader, more dynamic, and unpredictable.

\textbf{Future work.} Future research will focus on developing human-centered control strategies based on these empirical findings. First, we plan to recruit more diverse participants, including individuals with industrial backgrounds. Second, utilizing the interaction data collected on the unknown surface, we aim to extract implicit human compliance strategies to enhance the autonomous capability of the robot during contact-rich tasks in an unknown environment. Third, we will analyze how path geometry relates to operator states across the multi-segment path. This will suggest developing a geometry-aware adaptive control method to reduce operator effort. Finally, by integrating our proposed physical, cognitive, and autonomic workload metrics with environmental sensing, we aim to develop physiology-aware variable admittance control, which allows the robot to adapt compliance to both operator states and ambient environmental conditions.

\section{Conclusion}
\label{sec:conclusion}

This study evaluated operator state during contact-rich pHRI across varying environmental conditions. Our results reveal a decoupling between task performance and autonomic workload, underscoring the necessity of integrating multimodal physiological signals into human-centered robotic control.

\section*{Acknowledgment}
The authors would like to acknowledge the ﬁnancial support for this research received from the U.S. National Science Foundation (NSF) CMMI 2531678. Any opinions and ﬁndings in this paper are those of the authors and do not necessarily represent those of the NSF.

\bibliographystyle{IEEEtran}

\bibliography{ref}

\end{document}